\documentclass[sigconf]{acmart}
\AtBeginDocument{%
  }

\usepackage[capitalize]{cleveref}
\crefname{section}{Section}{Secs.}
\Crefname{section}{Section}{Sections}
\Crefname{table}{Table}{Tables}
\crefname{table}{Table}{Tabs.}

\usepackage{cleveref} 
\usepackage{multirow}
\newcommand{\et}[2]{${#1}^{\pm{#2}}$}
\newcommand{\etb}[2]{$\mathbf{{#1}}^{\pm{#2}}$}
\newcommand{\ets}[2]{$\textcolor{blue}{{#1}}^{\pm{#2}}$}
\usepackage{hyperref} 

\setcopyright{acmlicensed}
\copyrightyear{2025}
\acmYear{2018}
\acmDOI{3746027.3755401}
\acmConference[MM '25]{Proceedings of the 33rd ACM International Conference on Multimedia}{October 27--31,
  2025}{Dublin, Ireland}
\acmISBN{979-8-4007-2035-2/25/10}

\begin{document}

\title{Versatile Multimodal Controls for Expressive Talking Human}

\author{Zheng Qin}
\email{qinzheng@stu.xjtu.edu.cn}
\affiliation{
  \institution{National Key Laboratory of Human-Machine Hybrid Augmented Intelligence, Xi'an Jiaotong University, Ant Group}
  \city{Xi’an}
  \country{China}
}

\author{Ruobing Zheng}
\email{zhengruobing.zrb@antgroup.com}
\authornote{Co-first author. Project Lead.}
\affiliation{%
  \institution{Ant Group}
  \city{Beijing}
  \country{China}
}

\author{Yabing Wang}
\email{wyb7wyb7@stu.xjtu.edu.cn}
\affiliation{
\institution{National Key Laboratory of Human-Machine Hybrid Augmented Intelligence, Xi'an Jiaotong University}
\city{Xi’an}
\country{China}}

\author{Tianqi Li}
\email{shijian.ltq@antgroup.com}
\affiliation{%
  \institution{Ant Group}
  \city{Beijing}
  \country{China}
}

\author{Zixin Zhu}
\email{zixinzhu@buffalo.edu}
\affiliation{%
  \institution{University at Buffalo}
  \city{Buffalo}
  \country{USA}
}

\author{Sanping Zhou}
\email{spzhou@xjtu.edu.cn}
\affiliation{
  \institution{National Key Laboratory of Human-Machine Hybrid Augmented Intelligence, Xi'an Jiaotong University}
  \city{Xi’an}
  \country{China}
}

\author{Ming Yang}
\email{m-yang4 at u.northwestern.edu}
\affiliation{%
  \institution{Ant Group}
  \city{New York}
  \country{USA}
}

\author{Le Wang}
\email{lewang@xjtu.edu.cn}
\authornote{Corresponding author.}
\affiliation{%
  \institution{National Key Laboratory of Human-Machine Hybrid Augmented Intelligence, Xi'an Jiaotong University}
  \city{Xi’an}
  \country{China}
}


\begin{abstract}
In filmmaking, directors typically allow actors to perform freely based on the script before providing specific guidance on how to present key actions. AI-generated content faces similar requirements, where users not only need automatic generation of lip synchronization and basic gestures from audio input but also desire semantically accurate and expressive body movement that can be ``directly guided'' through text descriptions. Therefore,  we present VersaAnimator, a versatile framework that synthesizes expressive talking human videos from arbitrary portrait images. Specifically, we design a motion generator that produces basic rhythmic movements from audio input and supports text-prompt control for specific actions. The generated whole-body 3D motion tokens can animate portraits of various scales, producing talking heads, half-body gestures and even leg movements for whole-body images. Besides, we introduce a multi-modal controlled video diffusion that generates photorealistic videos, where speech signals govern lip synchronization, facial expressions, and head motions while body movements are guided by the 2D poses. Furthermore, we introduce a token2pose translator to smoothly map 3D motion tokens to 2D pose sequences. This design mitigates the stiffness resulting from direct 3D to 2D conversion and enhances the details of the generated body movements. Extensive experiments shows that VersaAnimator synthesizes lip-synced and identity-preserving videos while generating expressive and semantically meaningful whole-body motions.  \textcolor{pink}{\url{https://digital-avatar.github.io/ai/VersaAnimator/}}
\end{abstract}



\maketitle

\section{Introduction}
Recent advances in diffusion models have significantly inspired research in the domains of talking head~\cite{tian2025emo,wang2024v,xu2024vasa} and human animation~\cite{zhang2024mimicmotion,zhu2025champ,wang2024disco}, enhancing the quality and expressiveness of one-shot video generation. Recently, some innovative works~\cite{Corona_Zanfir_Bazavan_Kolotouros_Alldieck_Sminchisescu,lin2024cyberhost,meng2024echomimicv2,lin2025omnihuman} successfully attempt to integrate these two directions, achieving synchronized generation of facial expressions and body movements, evolving talking heads into talking humans.

\begin{figure*}
\centering
  \includegraphics[width=0.95\textwidth]{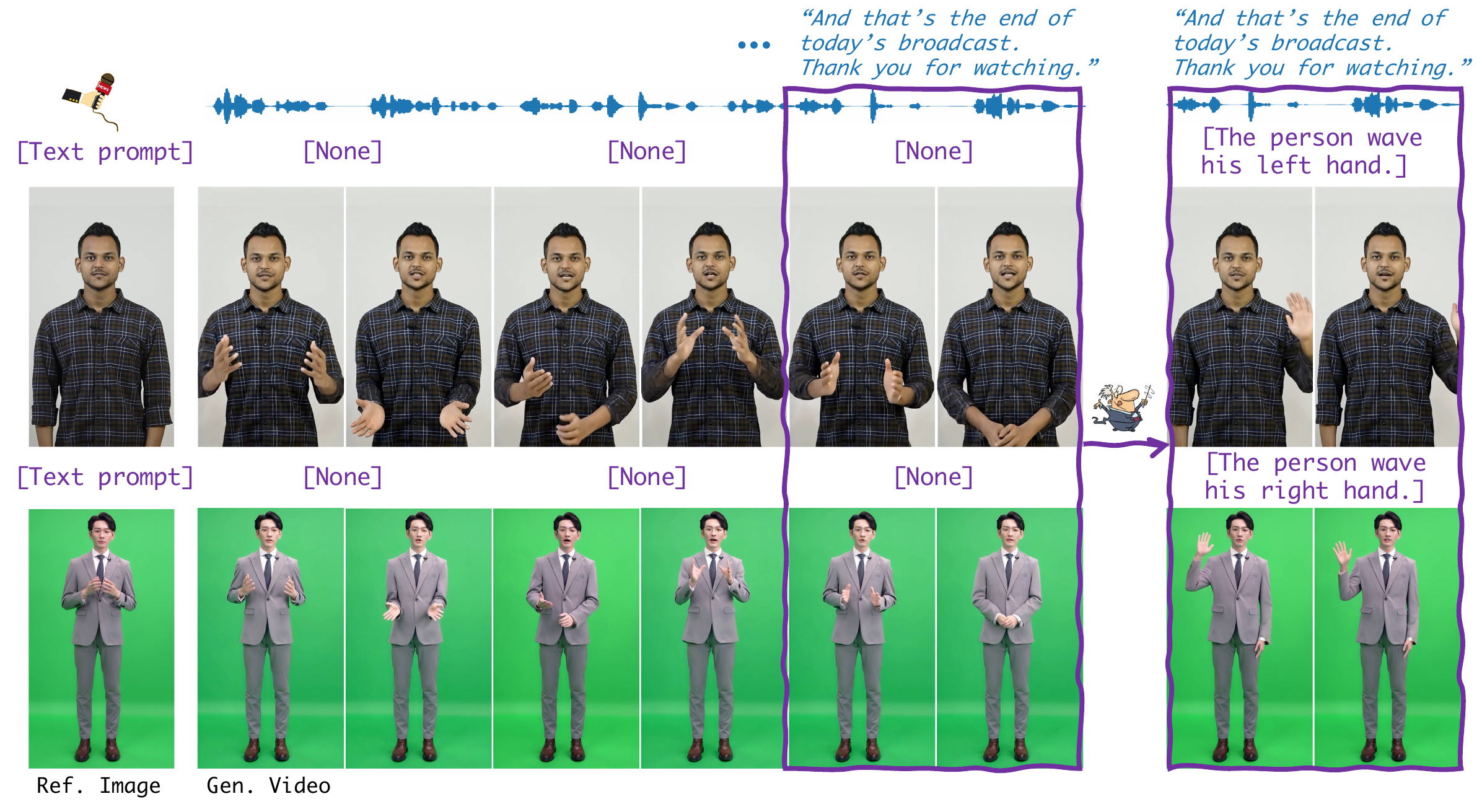}
  \caption{Given a reference image in the first column and an audio clip, our method generates photorealistic talking videos of the person. As demonstrated in the synthesized images, our approach supports arbitrary reference images, \textit{i.e.}, semi-body in the upper demo and whole-body in the lower, and allows users to control or edit the body motion by text prompts, \emph{e.g.} waving different hands at the closing.}
  \Description{Enjoying the baseball game from the third-base
  seats. Ichiro Suzuki preparing to bat.}
  \label{intro}
\end{figure*}

However, even state-of-the-art methods only demonstrate the potential to support some AI-generated content (AIGC) scenarios through demo videos. The quality of their generated videos and product functionality still struggle to meet the demands of commercial-grade scenarios. The limitations of existing methods primarily manifest in three aspects:


(1) Gestures generated from audio are simple, lacking expressiveness and semantic correspondence. Some methods generate random speech gestures based solely on audio, resulting in repetitive gestures with limited motion range that fail to convey sufficient semantic information. 

(2) Generated videos are not editable, including facial expressions and body movements. This leaves users with no direct means to adjust the results apart from regeneration, making outcomes entirely random. Achieving desired results becomes challenging and potentially time-consuming.

(3) The supported portrait scale is limited. Current methods mostly focus on gesture generation for half-body reference images, lacking the ability to generate corresponding whole-body movements, especially leg movements, for whole-body photographs. This limitation significantly constrains the range of producible video content.

Addressing these issues, we delve into their underlying causes and explore potential solutions. Unlike the direct correspondence between speech and lip movements, the relationship between body movements and speech content is a fuzzy many-to-many mapping. Certain specific, large-amplitude gestures are not entirely determined by speech alone but are also related to personal habits and contextual content. In film production, directors typically allow actors to perform freely based on the script before providing specific guidance on how to present key actions. AIGC faces similar requirements, where users not only need automatic generation of lip synchronization and basic gestures from audio input but also desire semantically accurate body movements that can be ``directly guided'' through text descriptions. Therefore, we argue that a reasonable motion generation model should  \textbf{use audio input to provide basic rhythmic movements, while expressive and semantically explicit actions should support user control through means such as text descriptions}. Additionally, we believe that movements should be generated based on whole-body motion representations. The range of motion generation should not be limited by the reference image, instead,  \textbf{the generated motions should be adaptable to and capable of driving reference images of any scale}.

Based on these considerations, we propose VersaAnimator, a versatile human animation framework that generates expressive talking human videos from arbitrary portrait images, not only driven by audio signals but also flexibly controlled by text prompts.  Specifically, we design a motion generator that produces basic rhythmic movements from the audio input and supports text-prompt control for specific actions. The generated whole-body 3D motion tokens can animate portraits of various scales, producing talking heads, half-body gestures and even leg movements for whole-body images. We utilize large-scale 3D motion datasets~\cite{human3d} to facilitate the learning of semantic associations between text descriptions and motion. A Vector Quantized Variational Autoencoder (VQ-VAE) is trained to unify motion representations across diverse datasets to a set of motion tokens. To enable multimodal control, we design a dual-branch transformer architecture that generates motion tokens conditioned on both audio signals and text prompts.  Besides, we introduce a multi-modal controlled video diffusion that generates photorealistic videos, where speech signals govern lip synchronization, facial expressions, and head motions while body movements are guided by the 2D poses. This diffusion model is trained in two stages, first learning the pose-to-video capability at different scales, and then the audio-driven facial motion generation. We also design a token-to-pose translator to smoothly map 3D motion tokens to 2D pose sequences. This design mitigates the stiffness resulting from direct 3D to 2D conversion and enhances the details of the generated body movements.

We have constructed a 40-hour human animation training set that spans from head to whole-body human videos. Extensive qualitative and quantitative experiments demonstrate that our method achieves state-of-the-art performance, synthesizing lip-synced and identity-preserving videos while generating expressive and semantically meaningful whole-body motions. Our work has the following main contributions:

\begin{itemize}
\item We propose a versatile talking human framework which, to our knowledge, is the first to simultaneously achieve audio-driven, text-controlled and whole-body scale motion generation, offering diverse application scenarios.
\item We propose a motion generation approach where audio controls basic motion rhythm and text controls semantically rich actions, and implement the generation of whole-body level 3D motion tokens conditioned on both audio signals and text prompts, capable of animating portrait images of various scales.
\item We design a multimodal video diffusion that can simultaneously control the fine-grained generation of facial expressions and body movements through audio and pose sequences, achieving high-quality and expressive human video generation.
\end{itemize}

\begin{figure*}[ht]
\begin{center}
\centerline{\includegraphics[width=\textwidth]{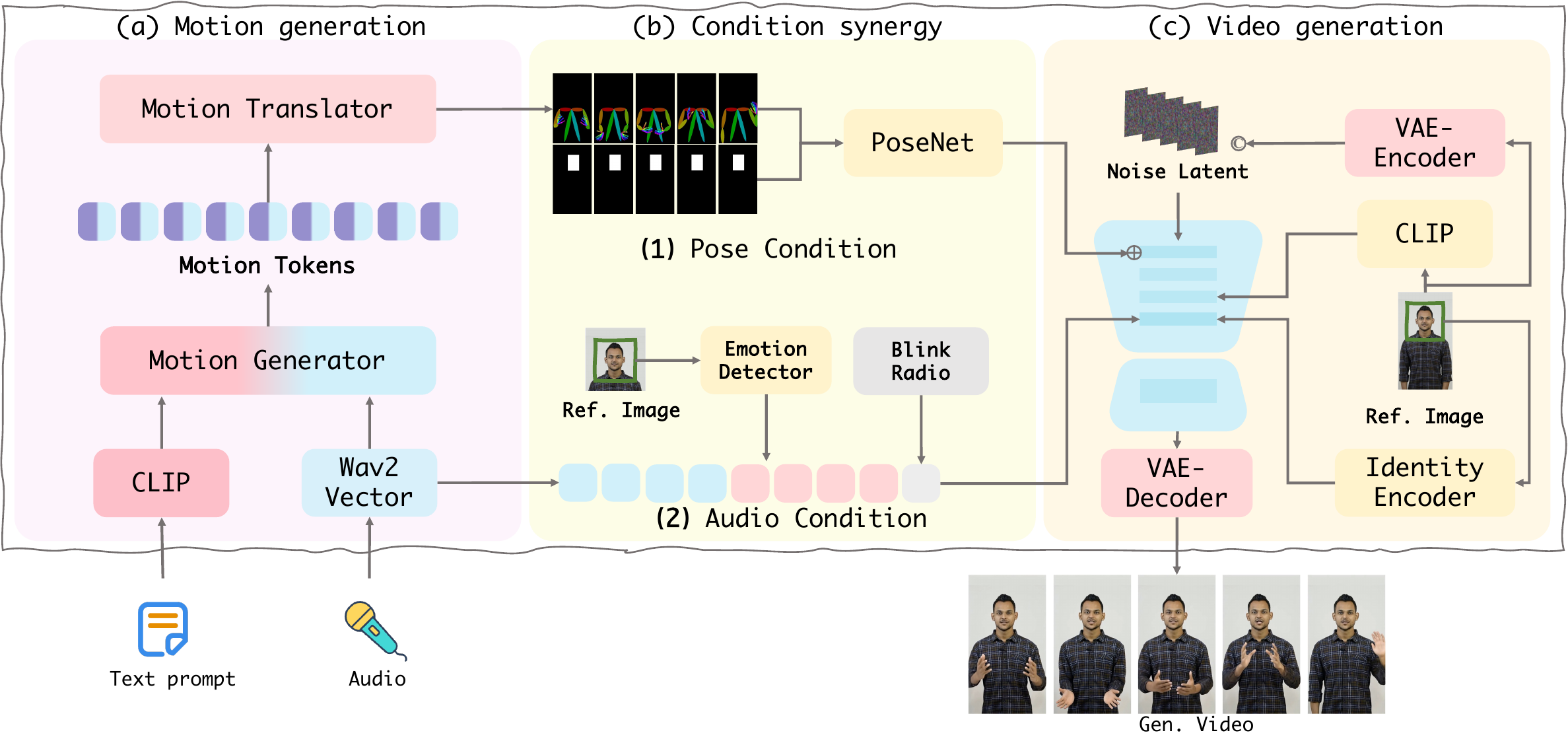}}
\caption{Overview of our VersaAnimator. The \textbf{Motion Generation} process uses both audio and text modalities to specify the facial and body motion to generate. Then \textbf{Condition Synergy} process generates the pose condition and audio condition as the control signal in video generation. Finally, we inject both conditions into the diffusion model to animate the reference character for \textbf{Video Generation}.}
\Description{}
\label{method1}
\end{center}
\end{figure*}

\section{Related work}
\textbf{Human Video Generation.}
In terms of the target region, existing works have focused on either head or body.
Remarkable efforts~\cite{tian2025emo,wang2024v,xu2024vasa,chen2024echomimic,zhang2023sadtalker,ma2023dreamtalk,wang2021one,yin2022styleheat,prajwal2020lip} have concentrated on audio-driven speaker video generation, primarily focusing on the head-shoulder region especially facial expressions, in speech-driven scenarios. 
Recent efforts~\cite{zhang2024mimicmotion,zhu2025champ,wang2024disco,feng2023dreamoving,xu2024magicanimate,hu2024animate,chang2023magicdance,karras2023dreampose,ma2024follow} have focused on driving character animation through pose guidance.
To enhance expressiveness and realism, VLOGGER~\cite{Corona_Zanfir_Bazavan_Kolotouros_Alldieck_Sminchisescu} and CyberHost~\cite{lin2024cyberhost} generate half-body talking videos with movements using only audio and reference maps as inputs.
EchoMimicV2~\cite{meng2024echomimicv2} need for gesture movements as input in addition to audio.
Despite recent advancements, they still exhibit several limitations, such as quite limited body motion and a lack of fine-grained control of gesture movements in the generated videos.

\textbf{Human Motion Generation.}
Human motion generation can be broadly categorized into two primary approaches \emph{w.r.t.} the control inputs: 
1) motion synthesis without conditions~\cite{raab2023modi, tevet2022human, Zhang_Black_Tang_2020} and 2) motion synthesis with specified multimodal conditions, such as action labels~\cite{xu2023actformer, lee2023multiact, dou2023c}, textual descriptions~\cite{wang2022humanise, chen2023executing, lu2023humantomato}, or audio and music~\cite{li2024lodge,tseng2023edge,li2022danceformer,  dabral2023mofusion}.
Due to its user-friendly nature and the convenience of language input, text-to-motion is one of the most important motion generation tasks. Given the remarkable success of diffusion-based generative models on AIGC tasks~\cite{rombach2022high}, some approaches have employed conditional diffusion models for human motion generation~\cite{zhang2024finemogen,chen2023executing}. Other works~\cite{zhang2023generating,guo2024momask,wang2023t2m} first discretize motions into tokens using vector quantization~\cite{van2017neural} and then predict the code sequence of motion.

\section{Method}
\label{Method}
\subsection{Preliminary and Overview}
\label{Preliminary}


\textbf{3D Human Motion Representation.}
SMPL (Skinned Multi-Person Linear Model)~\cite{loper2023smpl} is a widely used method for modeling human shape and posture, representing the 3D geometry of the human body through Linear Blend Skinning. 
It allows the body to adapt a variety of poses that are controlled and rationalized by a set of joint parameters, denoted as $j \in \mathbb{R}^{3J}$. 
Following HumanML3D~\cite{human3d}, a sequence of poses can express a continuous motion, represented as $\{j_t\}_{t=1}^T$. 
In particular, the 6D continuous rotation representation from \cite{zhou2019continuity} is developed to generate a compact yet comprehensive representation $\{ m_t\}_{t=1}^T$, where $m_t$ represents the motion representation at frame $t$, encompassing crucial details such as positional rotation and velocity. This approach proves advantageous for accurately modeling human motion.

3D human motions are highly controllable with a rich dataset available~\cite{human3d}, containing a variety of motion data paired with corresponding textual descriptions. 
This motivates us to utilize 3D motions as the whole-body motion representation.
On one hand, this enhances the authenticity and plausibility of human motions in videos. On the other hand, by leveraging the rich text-3D motion correspondence data, we can enable users to use convenient text prompts to control or edit the body motion generation flexibly in the generated video.

\textbf{Overview of VersaAnimator.}
Given an audio input and text prompts, our VersaAnimator can animate the reference character to synchronize with the speech while also following the textual prompt of body motions. 
The audio is divided into multiple clips for generating talking videos, and the text prompt controls the motion of the corresponding clip based on the user's specifications.
As shown in Figure~\ref{method1}, the inference process consists of three stages: (a) The Motion Generator~(\Cref{Audio-driven Motion Generation with Text Control}) takes both audio and text prompts as inputs, generating human motion in 3D conditioned on them, which is then input into the Motion Translator~(\Cref{code-pose translation construction}) to synthesize a 2D pose sequence. (b) After that, we construct the pose and audio conditions to be used in the next stage~(\Cref{Generating Photorealistic Talking and Moving Humans with Motion and Audio Control}). (c) We inject both conditions into the diffusion model to animate the reference character for video generation~(\Cref{Generating Photorealistic Talking and Moving Humans with Motion and Audio Control}). 

\subsection{Audio-driven Motion Generation with Text Control}
\label{Audio-driven Motion Generation with Text Control}
In this section, we first discretize the motions and then focus on generating motion tokens driven by both audio and text. To enable motion generation conditioned on multiple modalities, \textit{i.e.}, audio and text, we adopt a two-branch architecture. This architecture consists of a primary audio-to-motion branch and a plug-and-play text control branch, supported by a two-stage training strategy. 
\subsubsection{3D Human Motion Tokenizer} $ $

Vector Quantized Variational Autoencoders (VQ-VAE)~\cite{van2017neural} enable the learning of discrete representations, offering significant advantages for content compression and generation. VQ-VAE reconstructs the motion sequence using an autoencoder with a $K$-size learnable  codebook $C = \{c_k\}_{k=1}^K$, where $c_k \in \mathbb{R}^{d_c}$ represents the discretized motion token and $d_c$ denotes the feature dimension.
Given a motion sequence $M = \{m_t\}_{t=1}^T$, the encoder $\mathcal{E}$ maps it into a latent feature sequence $Z = \{z_i\}_{i=1}^{T/l}$, where $z_i \in \mathbb{R}^{d_c}$ and $l$ is the temporal downsampling rate of $\mathcal{E}$. For each latent feature $z_i$, quantization is performed by selecting the closest element in $C$, resulting in the quantized feature sequence $\hat{Z} = \{\hat{z}_i\}_{i=1}^{T/l}$, as follows:
\[
\hat{z}_i = \underset{c_k \in C}{\arg \min} \left\|z_i - c_k\right\|_2,
\]
where $\hat{z}_i$ is the quantized version of $z_i$. Finally, the decoder $\mathcal{D}$ reconstructs the motion sequence $\hat{M}$ from the quantized feature sequence $\hat{Z}$. We construct a 512-size codebook, where the motion tokens are used to combine and express coherent movements of the human body.

\subsubsection{Motion Generation Architecture} $ $
\label{Motion Generation under Noise and Text Conditions}

\textbf{Audio Token.} 
We utilize wav2vec~\cite{schneider2019wav2vec} as our audio tokenizer and we design a temporal block consists of multiple convolution layers to extract and fuse the temporal information.

We implement the transformer $p_\theta^{audio}(\hat{z} | audio)$ as the primary audio-to-motion architecture, where $\hat{z}$ denotes the predicted motion token here. The goal is to predict the code sequence~(responded to motion tokens in codebook) conditioned by the audio signal. 
Specifically, the audio tokens are fed into the transformer encoder to capture long-range dependencies and contextual relationships within the audio sequence. The outputs of each transformer layer are gathered and denoted by $\{f_s^{audio}\}_{s=1}^S$, where $f^{audio}_s$ represent the $s$-th layer's output and $S$ is set to 8. 
Then utilizing simple linear layers, we convert last feature $f_S^{audio}$ to code probability $I^\text{audio} \in \mathbb{R}^{T \times K}$, where $T$ and $K$ represent the predicted motion length and size of the codebook respectively. At timestamp $t$, the probability distribution of the motion tokens in codebook is as follows:
\begin{equation}
\begin{aligned}
P(c_k) = p_k^t ,  \ \  k = 1,...,K. \\
\end{aligned}
\end{equation}
We then select the code sequence that aligns with the audio signal based on the probability distribution $I^{audio}$, and use the codebook to retrieve the corresponding motion tokens $\{\hat{z}_i\}_{i=1}^{T/l}$.
Finally, a trained decoder converts them into the motion sequence ${\hat{M}} = \{\hat{m}_t\}_{t=1}^T$.

\begin{figure}[t]
\begin{center}
\centerline{\includegraphics[width=\columnwidth]{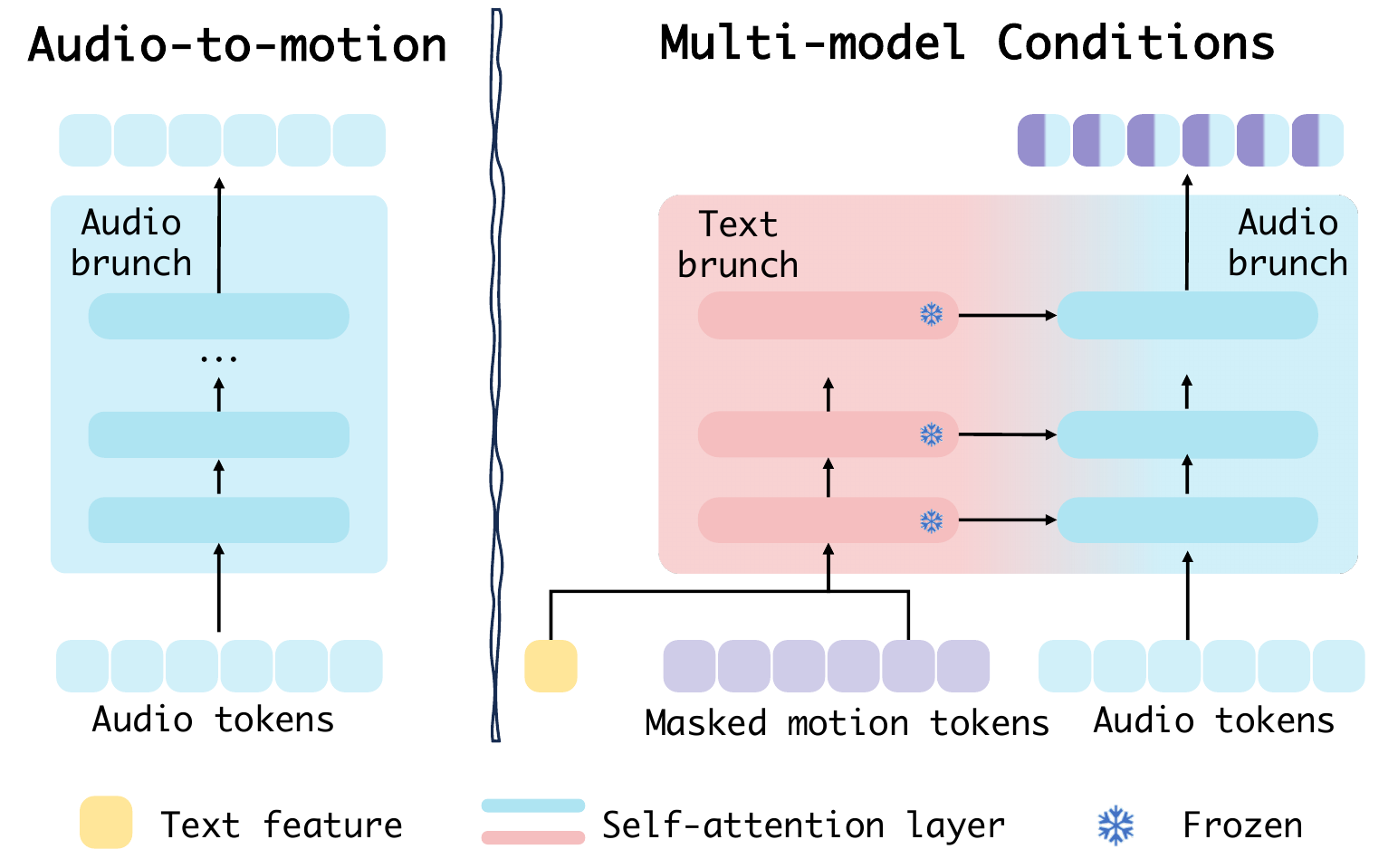}}
\caption{Structure of the motion generator. Left: The primary audio-to-motion architecture. Right: The two-branch transformer that generates motions conditioned on both audio and text prompts.}
\Description{}
\label{method2}
\end{center}
\end{figure}

\textbf{Multimodel Conditions.}
To edit the motion of animated character with text, \textit{i.e.}, to allow the text to control the prediction of the motion token, we adopted a two-branch transformer $p_\theta^{}(\hat{z} | text, audio)$. Specifically, we learn a text-to-motion transformer to model the tokens conditioned on text. The text-to-motion transformer has the same architecture as the primary audio-to-motion model. Given the text signal, we use CLIP~\cite{radford2021learning} for extracting text features and feed into text-to-motion transformer and obtain $\{f_s^{text}\}_{s=1}^S$, where $f^{text}_s$ represents the $s$-th layer's output. We then consolidate two transformer structure by summing them layer by layer as show in~\Cref{method2} .

\subsubsection{Training} $ $

\textbf{Training 3D Human Motion Tokenizer.} 
Overall, the VQ-VAE is trained via a motion reconstruction loss combined with a latent embedding loss at quantization layer:
\begin{equation}
\begin{aligned}
\mathcal{L}_{m d r}=\|M-\hat{M}\|_1+\beta \left\|Z-\operatorname{sg}\left[\hat{Z}\right]\right\|_2^2 ,
\end{aligned}
\end{equation}
where $\operatorname{sg}[\cdot]$ denotes the stop-gradient operation, and $\beta$ a weighting factor for embedding constraint. The codebooks are updated via exponential moving average and codebook reset following~\cite{zhang2023generating}.

\textbf{Training Motion Generation Model.} 
We first train the text to motion branch, then freeze the weights and train the audio to motion branch.
For text to motion branch, we randomly mask out the sequence elements, by replacing the tokens with a special mask token. The goal is to predict the masked tokens given text. We use CLIP~\cite{radford2021learning} for extracting text features. 
Our training goal is to predict the masked tockens. 
We directly maximize the log-likelihood of the data distribution $p(\hat{Z} \mid text)$:
\begin{equation}
\begin{aligned}
\mathcal{L}_{\text {tran }}=\mathbb{E}_{\hat{Z} \sim p(\hat{Z})}[-\log p(\hat{Z} \mid text)],
\end{aligned}
\end{equation}
the likelihood of the full sequence is denoted as follows: 
\begin{equation}
\begin{aligned}
p(\hat{Z} \mid text)=\prod_{i=1}^{T/l} \left(p\left(\hat{z}_t \mid text\right) \cdot \left(1 - [mask]_t\right) + [mask]_t\right),
\end{aligned}
\end{equation}
where $[mask]_t$ indicates whether the $i$-th token is masked, which is set to 0 if masked, and 1 otherwise.

During training the audio-to-motion branch, we set the text ``A person is giving a speech.'' as text control. Given the input text control and audio tokens, the training objective is:
\begin{equation}
\begin{aligned}
\mathcal{L}_{\text {audo }}= \sum_{i=1}^n-\log p_\phi\left(\hat{z}_i \mid a_i, text\right),
\end{aligned}
\end{equation}
where $\hat{z}_i$ and $a_i$ are the $i$-th motion and audio token respectively.

\subsubsection{Token2pose Translator Construction} $ $
\label{code-pose translation construction}


After obtaining the generated motion tokens, the next step is to map them to 2D pose sequences for video generation. Directly mapping 3D human model to 2D pose often results in stiff and unrealistic motions. To address this issue, we construct a relation bank that links all motion tokens to detailed 2D poses.  Specifically, we collect whole-body template videos covering various types of actions, simultaneously extracting normalized 2D pose and SMPL-X data. Next, we use the previously trained 3D human motion tokenizer to convert SMPL-X data into motion tokens. By extracting real-world poses from template videos and aligning them with motion tokens,  we can create a large set of token2pose pairs. We use these pairs to map the generated motion tokens to real 2D poses and further adapt them to the target identity, thereby enhancing the realism of the generated motions.

\subsection{Generating Photorealistic Talking and Moving Humans with Audio and Motion Control}
\label{Generating Photorealistic Talking and Moving Humans with Motion and Audio Control}
We aim to generate videos with accurate lip synchronization and rich gestures from a single image and audio input. Since there is only indirect weak correlation between audio and body movements, generating whole-body movements solely based on audio input remains challenging. Therefore, we pre-generate motion representations using the aforementioned Motion Generation module, then transform these motion representations into explicit pose sequences, combined with audio and a single reference image to produce the final video. To accomplish this task, we first construct a multimodal-controlled video diffusion model.

\begin{figure}[t]
\begin{center}
\centerline{\includegraphics[width=\columnwidth]{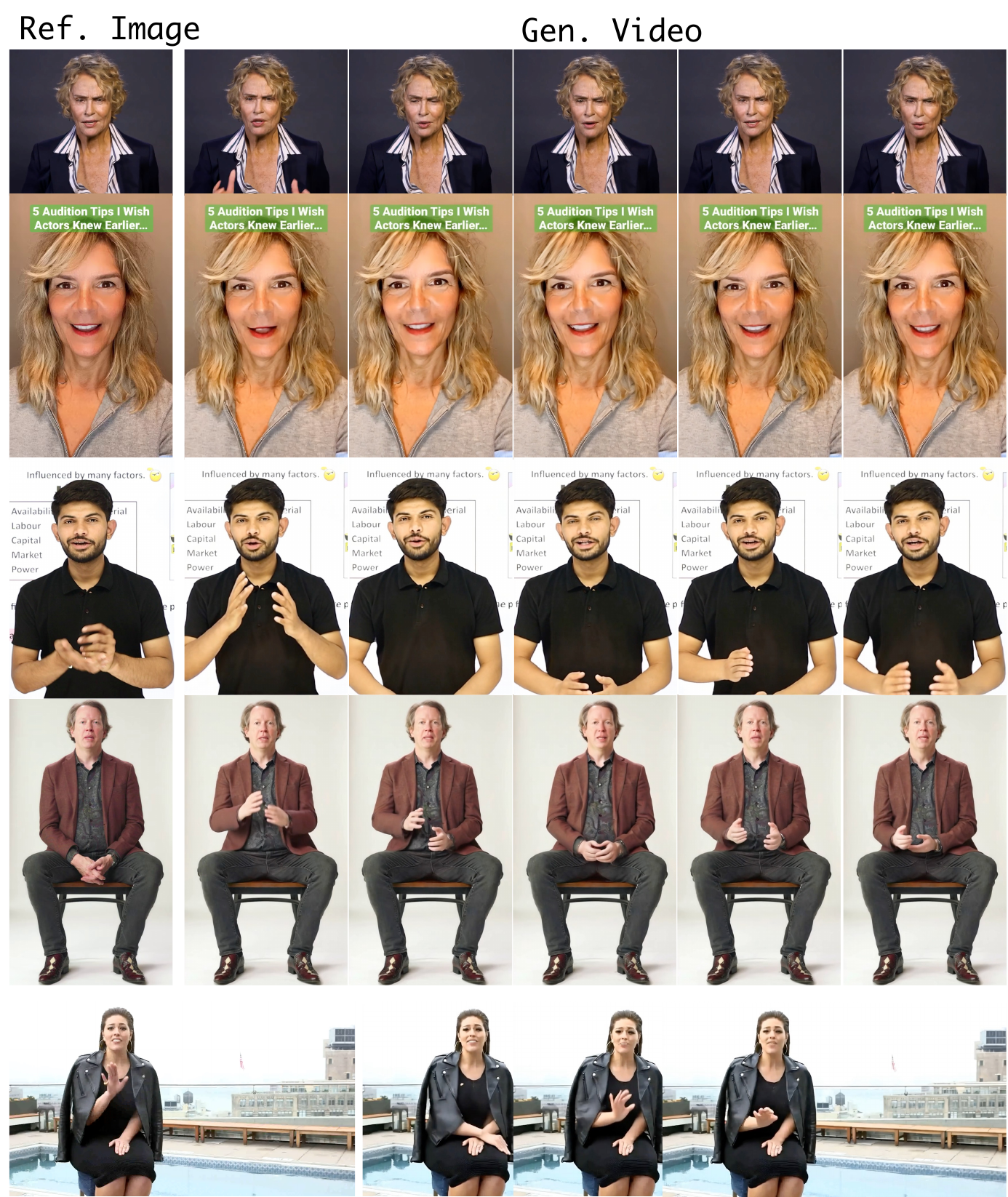}}
\caption{Results on Multi-Animate with different audio and reference images (ranging from head to whole-body).}
\Description{}
\label{exp1}
\end{center}
\end{figure}

\begin{figure}[t]
\begin{center}
\centerline{\includegraphics[width=\columnwidth]{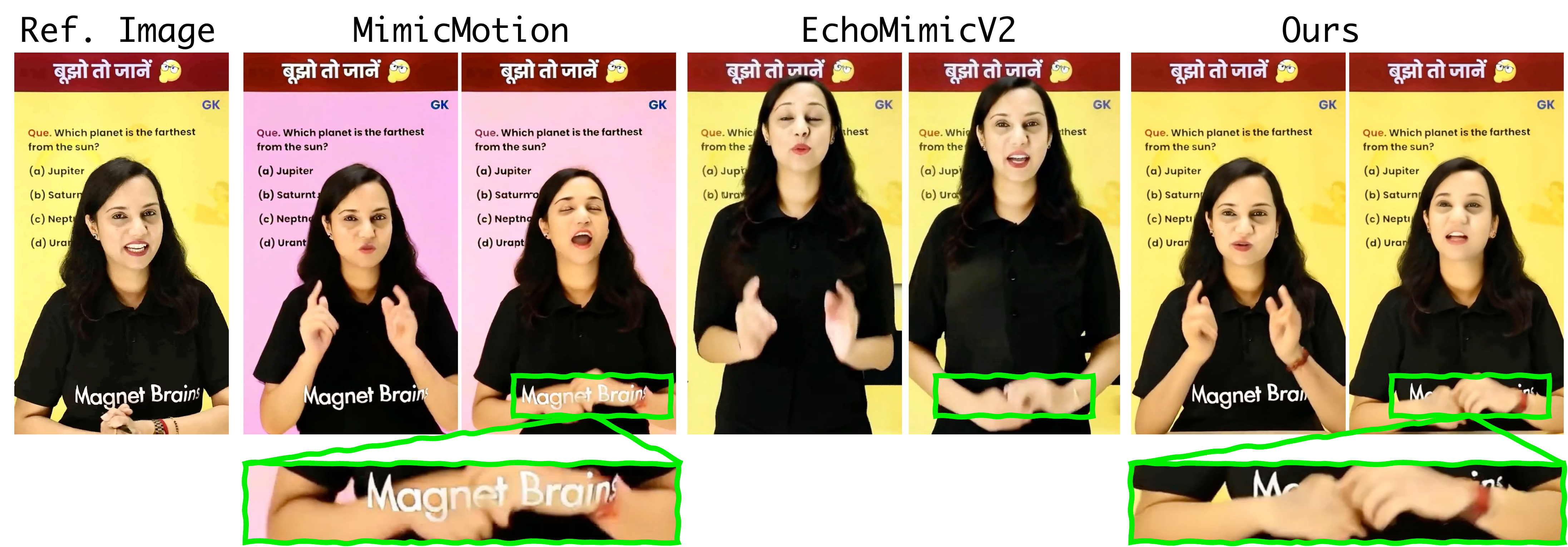}}
\caption{Comparisons of Detail Preservation. Focus on the background preservation, hand clarity, the occlusion relationship between the hands and the clothing pattern.}
\Description{}
\label{exp6}
\end{center}
\end{figure}

\begin{figure*}[ht]
\begin{center}
\centerline{\includegraphics[width=0.85\textwidth]{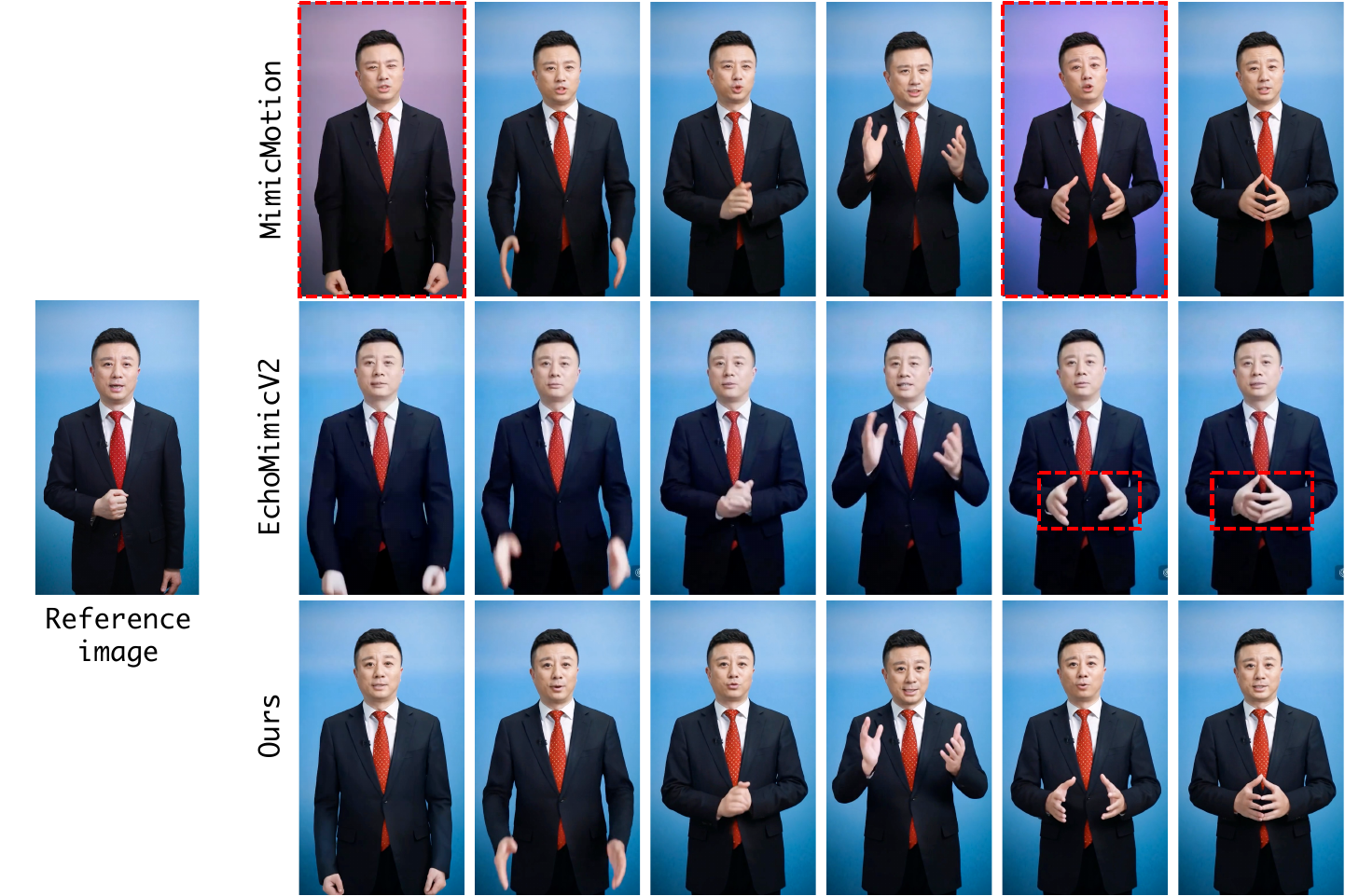}}
\caption{Comparisons with pose-driven body animation methods. The red dashed boxes indicate areas with poor performance, such as background changes or unclear hands.}
\Description{}
\label{exp7}
\end{center}
\end{figure*}

\subsubsection{Co-Speech Human Animation with Video Diffusion Model} $ $

We extend an off-the-shelf human animation framework~\cite{zhang2024mimicmotion} by incorporating additional conditional signal, including audio, emotion labels, and blinking ratios. This enhancement enables the simultaneous generation of lip movements and whole-body animations.

As shown in~\Cref{method1}, we refine the original whole-body pose sequence extracted from video frames~\cite{hu2024animate} into a composite representation, which combines the body pose below the neck with a fixed-size head mask centered on the facial midpoint above the neck. This design circumvents the impact of facial keypoints in original pose sequences on generating synthesized facial movements, thereby enabling lip movements and facial expressions in generated videos to be driven by subsequent audio inputs and other control signals. This composite representation is then mapped through PoseNet and element-wise added to the output of the U-Net's first convolution layer~\cite{zhang2024mimicmotion}. In addition, along with extracting image features from the reference image, we detect the facial region and extract an identity embedding using a pre-trained face recognition network~\cite{deng2019arcface}. We draw inspiration from the Cyberhost~\cite{lin2024cyberhost} and introduce an additional cross-attention layer after the original cross-attention layer in every U-Net block, specifically designed to generate facial dynamics. Notably, we adopt the strategy from ~\cite{li2024ditto} to incorporate conditional signal directly related to facial motion, including audio, expression labels, and blink ratio, as an additional control signal that interacts with the identity embedding in the new cross-attention layer. This design mitigates the ambiguity inherent in purely audio-driven approaches, leading to accurate generation of lip motions, facial expressions, and eye blinking while maintaining strong identity consistency.

\subsubsection{Training Strategy} $ $

We employ a two-stage training strategy to train the rendering module. Given that the original pose-guided SVD backbone~\cite{zhang2024mimicmotion} is typically trained on whole-body dance data, we first fine-tune it using collected close-up human talking videos to enhance the generation of magnified facial details. Subsequently, we introduce additional control signals and train the complete network controlled by both audio and pose. In the second stage, we utilize human talking videos with audio-visual synchronization at various scales, including whole-body standing postures, close-up seated postures, and the talking-head scale. For each training videos, we extract audio features as ~\cite{xu2024hallo} and randomly select one frame as a reference image to extract image features and identity features~\cite{lin2024cyberhost}.

\section{Experiments}
\label{Experiments}
\subsection{Setting}
\textbf{Training Data and Evaluation Benchmark Contraction.}
To train the multimodal diffusion model, we collect about 40 hours of human talking videos with a variety of visible body regions, from head to whole body, including multiple nationalities, languages, and ages. We sample from the above videos to create an audio-to-motion dataset and combine it with the HumanML3D~\cite{human3d} dataset to train a text-controlled, audio-driven motion generator.

Publicly available datasets typically focus on evaluating audio-driven talking head or pose-driven character animations, while whole-body talking videos are scarce in existing datasets.
To evaluate human animation comprehensively from multiple perspectives such as facial generation, motion generation, and rendering quality, we introduce a multi-scale human animation evaluation benchmark, named Multi-Animate. 
This benchmark features human talking videos ranging from head-only to whole-body animations. Our dataset includes 30 human talking videos with 100 speech segments, covering various body scales, nationalities, and languages.

\begin{table*}[t]
\begin{center}
\caption{Quantitative comparison and ablation  study of our proposed VersaAnimator on multi-scale animation benchmark, \textit{Multi-Animate.} Bolding indicates the best result among state-of-the-art methods.}
\label{table1}
\resizebox{0.85\textwidth}{!}{
\begin{tabular}{lcccccccc}
\hline \hline
Methods              & FID$\downarrow$ & FVD$\downarrow$ & SSIM$\uparrow$ & PSNR$\uparrow$ & E-FID$\downarrow$ & Sync-D$\downarrow$ & Sync-C$\uparrow$ & CSIM$\uparrow$ \\ \hline
MimicMotion~\cite{zhang2024mimicmotion}          & 39.25           & 220.69          & 0.13           & 16.64          & \textbf{1.90}              & 9.97               & 4.96             & 0.65           \\
EchoMimicV2~\cite{meng2024echomimicv2}          & 38.40            & 443.48          & 0.19           & 15.51          & 2.92              & 9.46               & 5.34             & 0.63           \\
VersaAnimator        & \textbf{16.28}           & \textbf{128.40}          & \textbf{0.26}  & \textbf{20.69} & 3.12              & \textbf{8.85}               & \textbf{6.55}             & \textbf{0.99}           \\ \hline
VersaAnimator*       & 29.28           & 235.84          & 0.32           & 16.44          & 7.73              & 8.57               & 6.75             & 0.74           \\ 
w/o token2pose translator   & 32.34           & 302.87          & 0.23           & 14.31          & 8.42              & 8.23               & 6.91             & 0.73           \\
w/o per-layer fusion & 30.14           & 240.56          & 0.30           & 16.12          & 7.23              & 8.59               & 6.64             & 0.79           \\
w/o text branch      & 29.12           & 220.67          & 0.34           & 17.02          & 7.45              & 8.34               & 6.78             & 0.71           \\ \hline \hline
\end{tabular}
}
\end{center}
\end{table*}

\begin{table}[t]
\centering
\caption{Evaluation of text-to-motion capability on the HumanML3D test set, with comparison to state-of-the-art methods such as TM2T~\cite{guo2022tm2t}, T2M~\cite{guo2022generating}, MDM~\cite{shafir2023human}, MLD~\cite{chen2023executing}, MotionDiffuse~\cite{zhang2022motiondiffuse}, T2M-GPT~\cite{zhang2023generating}, and ReMoDiffuse~\cite{zhang2023remodiffuse}.}
\label{table2}
\resizebox{0.9\linewidth}{!}{
\begin{tabular}{lcccc}
\hline\hline
\multirow{2}{*}{{Methods}}           & \multicolumn{3}{c}{{R Precision$\uparrow$}}       & \multirow{2}{*}{{FID$\downarrow$}} \\ \cline{2-4}
                                            & {Top 1}   & {Top 2}    & {Top 3}    &                                  \\ \hline
TM2T                  & \et{0.424}{.003} & \et{0.618}{.003}  & \et{0.729}{.002}  & \et{1.501}{.017}                          \\
T2M    & \et{0.455}{.003} & \et{0.636}{.003}  & \et{0.736}{.002}  & \et{1.087}{.021}                          \\
MDM                  & -                & -                 & \et{0.611}{.007}  & \et{0.544}{.044}                          \\
MLD        & \et{0.481}{.003} & \et{0.673}{.003}  & \et{0.772}{.002}  & \et{0.473}{.013}                          \\
MotionDiffuse & \et{0.491}{.001} & \et{0.681}{.001}  & \et{0.782}{.001}  & \et{0.630}{.001}                          \\
T2M-GPT        & \et{0.492}{.003} & \et{0.679}{.002}  & \et{0.775}{.002}  & \et{0.141}{.005}                          \\
ReMoDiffuse    & \etb{0.510}{.005} & \ets{0.698}{.006}  & \etb{0.795}{.004} & \ets{0.103}{.004}                          \\ \hline
Ours                                        & \ets{0.505}{.002} & \etb{0.698}{.003} & \ets{0.794}{.002}  & \etb{0.090}{.003}                         \\ \hline\hline
\end{tabular}
}
\end{table}

\textbf{Metrics.}
We evaluate the performance of text-to-motion generation using R-Precision~\cite{guo2022generating} and Frechet InceI’m ption Distance (FID)~\cite{heusel2017gans}, which reflect semantic consistency and the overall motion
quality, respectively.
For human video generation, we employ a range of metrics.
FID, FVD~\cite{unterthiner2018towards}, SSIM~\cite{wang2004image}, and PSNR~\cite{hore2010image} are used to assess low-level visual quality, while E-FID~\cite{deng2019accurate} is used to evaluate the authenticity of the generated images. CSIM is employed to measure identity consistency. Additionally, we use SyncNet~\cite{prajwal2020lip} to calculate Sync-C and Sync-D, which validate the accuracy of audio-lip synchronization.

\textbf{Implementation Details.}
We implemented our VersaAnimator in PyTorch~\cite{paszke2019pytorch}, and performed all experiments on NVIDIA A100 GPUs (80GB).
The motion generation and video generation components are trained on 1 and 8 GPUs, respectively. All implementation and training details are provided in the supplementary material.

\subsection{Qualitative Results}
\textbf{Multi-scale Human Animation.}
Given an audio, our method generates a talking video featuring the character from the reference image. As shown in \Cref{exp1}, we present the generated results on Multi-Animate. The samples we tested cover head, half-body, and whole-body scales, all of which produce coherent, realistic, and natural speaker videos. 
As shown in the bottom row, our method does not require restricting the character's position; it can appear anywhere in the reference image.
These results demonstrate that our proposed VersaAnimator generalizes effectively to diverse characters and body scales.

\begin{figure}[t]
\begin{center}
\centerline{\includegraphics[width=0.9\columnwidth]{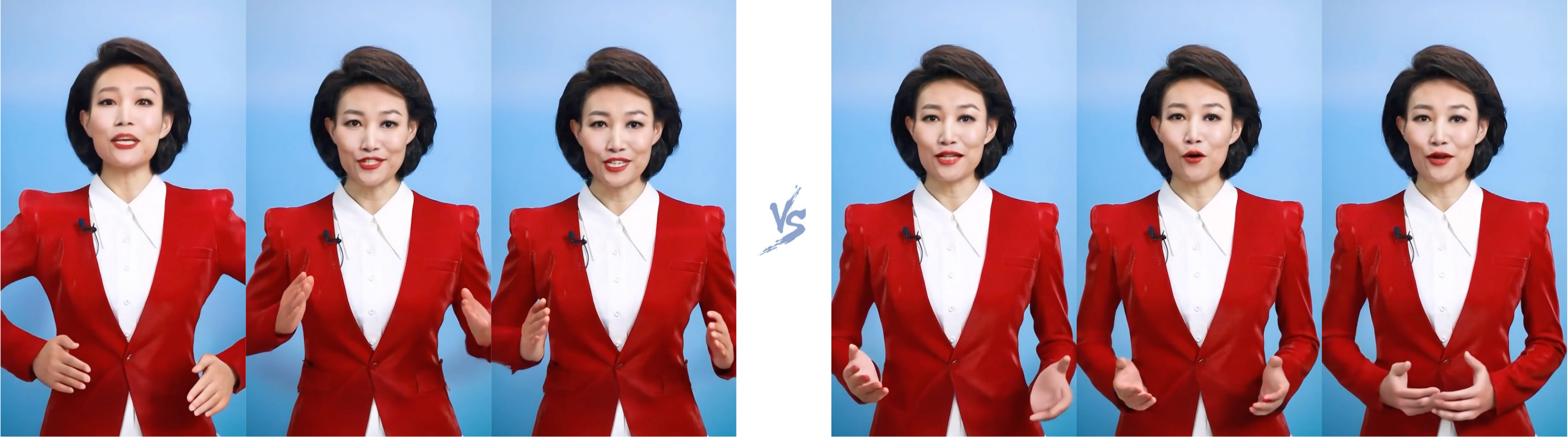}}
\caption{Qualitative comparison on whether to use token2pose translator. Left: Stiff and unnatural without pose translation. Right: More natural and fluid motion with pose translation.}
\Description{}
\label{exp3}
\end{center}
\end{figure}

\begin{figure}[t]
\begin{center}
\centerline{\includegraphics[width=\columnwidth]{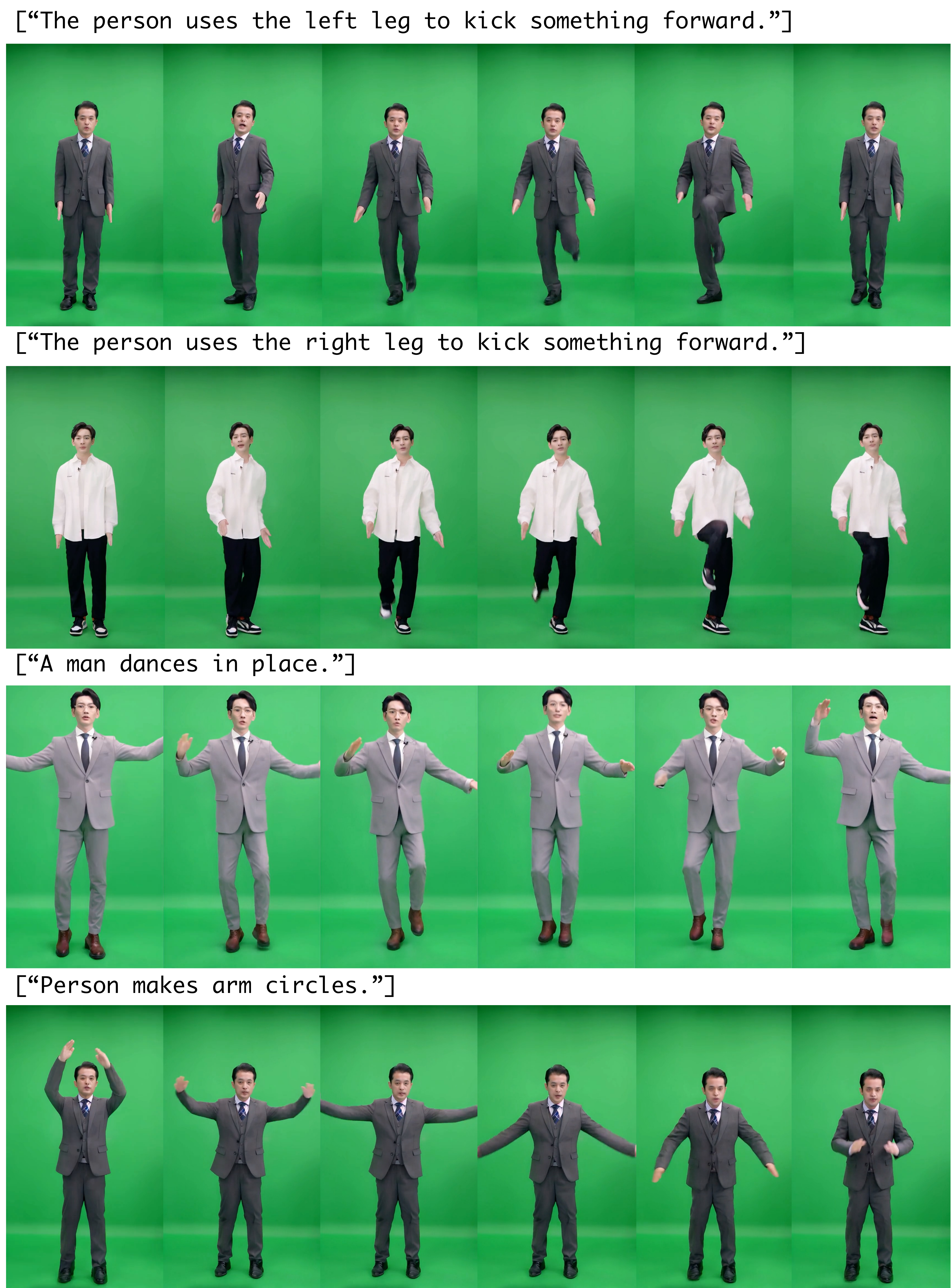}}
\caption{Visual illustration of text control for customizing the character's motion in the generated video.}
\Description{}
\label{exp2}
\end{center}
\end{figure}

\textbf{Comparisons of Detail Preservation.}
As shown in \Cref{exp6}, our method preserves fine details from the reference image, such as clothing patterns, background, and character position. Notably, when hands overlap with clothing (green box), the occluded regions are not rendered, consistent with physical laws.

\textbf{Comparisons with Pose-Driven Body Methods.}
\Cref{exp7} also shows that VersaAnimator achieves better clarity in local areas, such as the hands, with additional comparisons provided in the supplementary material.


\textbf{Visual Illustration of Text Control.}
As shown in ~\Cref{exp2}, we present several cases of text prompts to customize the character's motion in the video.
These motions are absent from the training videos. This demonstrates that our method can enhance the diversity of generated actions in videos and improve user control over the video generation process, which will improve the diversity of movements in some entertainment scenarios such as stand-up comedy, and can support the user's verbal intervention in video generation.

\subsection{Quantitative Results}

\textbf{Multi-scale Talking Body Evaluation.} 
To evaluate the performance of human video generation, we conduct a comprehensive comparison with the current open-source methods, MimicMotion and EchoMimicV2. Our input also includes the pose for alignment.
As shown in~\Cref{table1}, our VersaAnimator outperforms the state-of-the-art methods across most key metrics, ranking first in quality metrics (FID, FVD, SSIM, and PSNR), synchronization metrics (Sync-C and Sync-D), and the consistency metric (CSIM).
* indicates the version where only audio is used as input, which is more difficult but more valuable. 
Even in these challenging settings, the purely audio-driven version still performs on par with other methods that use pose prompts.
Note that our benchmark includes talking videos with a variety of visible body regions. These results demonstrate VersaAnimator's robustness and strength.

\textbf{Text-to-motion Evaluation.}
In order to test the text control capabilities in our pipeline, we keep the audio during evaluation, and we randomly sample audio pieces as input for each test sample. 
\Cref{table2} presents a comparison of our method with state-of-the-art approaches. Our VersaAnimator outperforms these methods in most key metrics, ranking first in R-Precision (Top-2) and FID, and second in R-Precision (Top-1, Top-3). This demonstrates that, even with audio as an additional input, the motions we generate maintain strong semantic consistency and realism.

\subsection{Ablation Study}
\textbf{Analysis of Token2pose Translator.}
As shown in~\Cref{table1}, row 5 indicates the 
impact of our Token2pose Translator. It significantly enhances the quality metric~(FID, FVD, SSIM, and PSNR). As shown in~\Cref{exp3}, this module largely alleviates the stiffness and unreality associated with 3D human modeling.

\textbf{Analysis of the Fusion Method for Text and Audio Conditions.}
To evaluate the multimodal fusion strategy, we replace the per-layer fusion approach with fusion at the last layer. As shown in \Cref{table1}, row 6 demonstrates that this strategy significantly improves all metrics by progressively fusing audio and text clues at each layer, thereby enhancing the accuracy of the generated output. 
To verify the impact of the text branch on the original generative capability of the audio, we remove the text branch. As shown in row 7, this fusion strategy allows both modalities to collaborate effectively without audio and text clues.


\section{Conclusion}
\label{Conclusion}
In this paper, we highlight key challenges in human animation, particularly the need to support diverse scenarios involving varying portrait scales and the fine-grained control of body motion through text prompts. We introduce VersaAnimator, a versatile talking human framework that generates natural and expressive videos from static portraits. VersaAnimator can animate portraits of various scales while allowing users to customize body movements, all with well-synchronized lip and facial expressions. Extensive experiments demonstrate that VersaAnimator outperforms existing methods, offering more intuitive user interaction and significantly enhancing the overall human animation experience.

\section{Acknowledgments}
This work was supported in part by the National Key Research and Development Project under Grant 2024YFB4708100, National Natural Science Foundation of China under Grants 62088102, U24A20325 and 12326608, Key Research and Development Plan of Shaanxi
Province under Grant 2024PT-ZCK-80 and Ant Group Research Intern Program.

\bibliographystyle{ACM-Reference-Format}
\balance
\bibliography{sample-base}

\newpage
\appendix
\title{Versatile Multimodal Controls for Expressive Talking Human Animation}

\section{Implementation Details.}
For text-controlled, audio-driven motion generation, we set the codebook as 512 × 512, \textit{i.e.}, 512 512-dimension dictionary vectors. Following~\cite{guo2024momask,zhang2023generating}, the dataset HumanML3D is extracted into motion features with dimensions 263, which related to local joints position, velocity, and rotations in root space as well as global translation and rotations. The joint number is set to 22. The transformer is composed of 8 transformer layers, with 6 heads and a latent dimension of 384.

We train our diffusion model on 40 hours of self-recorded and web-collected news broadcast videos with synchronized audio and visual content. The dataset is balanced across diverse ethnicities, languages, and shot scales. The average length of the training video clips is 10 seconds. The video diffusion module is trained on 8 NVIDIA A100 GPUs (80GB). The first stage, pose2vid, is trained for 10,000 steps, building upon pre-trained weights from Mimicmotion~\cite{zhang2024mimicmotion}. The second stage, which incorporates audio input, is trained for an additional 36,000 steps.

\section{Details of Audio Token.}
To enhance the encoding of audio for motion-driven animation, we utilized wav2vec~\cite{schneider2019wav2vec} as our audio feature encoder. Specifically, we concatenated the audio embeddings from the final 12 layers of the wav2vec model to capture a richer and more diverse range of semantic information across different audio layers. Considering the sequential nature of audio data and its contextual dependencies, we designed a temporal block to extract and fuse the temporal information. Through multiple convolution layers, we transformed the pre-trained audio embeddings into $\left\{c_{\text {audio}}^{t}\right\}_{t=1}^{T/l}$, where $c_{\text {audio }}^{t} \in \mathbb{R}^{D_a}$ represents the $t$-th audio token. Here, $l$ is the downsampling rate of the temporal block and ${D_a}$ denotes the dimension of the audio token.

\section{Comparison of Motion Amplitude.}
\begin{figure*}[ht]
\vskip 0.2in
\begin{center}
\centerline{\includegraphics[width=\textwidth]{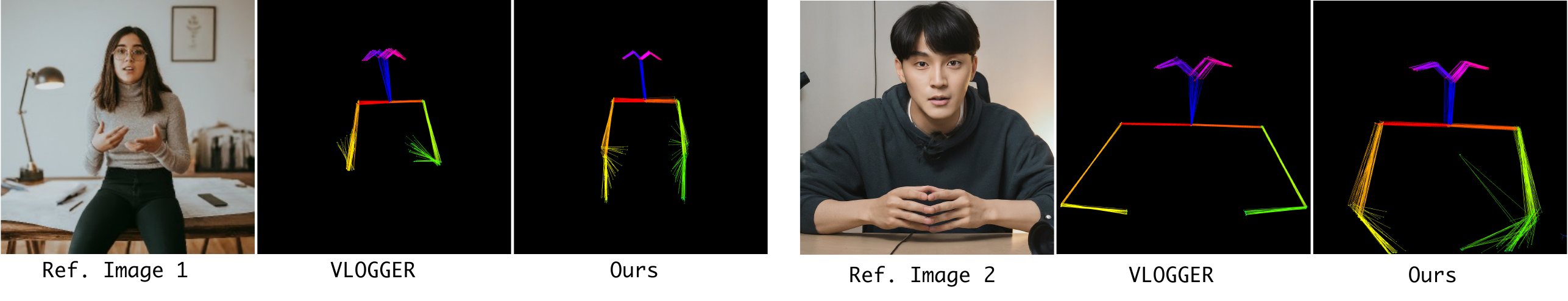}}
\caption{Comparison of motion amplitude with VLOGGER~\cite{Corona_Zanfir_Bazavan_Kolotouros_Alldieck_Sminchisescu}. We extract the dwpose frame by frame from the generated video and visualize the dwpose of the upper limbs. The dwpose of all the frames is superimposed on a single image, representing the entire video. It can be seen that our method covers a larger active area and is more expressive.}
\Description{}
\label{exp4}
\end{center}
\vskip -0.2in
\end{figure*}

We observe that the motion generated by other methods exhibits a limited magnitude around the gesture in the reference image. We download several videos from the official VLOGGER website and test our method for comparison. We extract the dwpose frame by frame from the generated video and visualize the dwpose of the upper limbs. The dwpose of all the frames is superimposed on a single image, representing the entire video. As shown in~\Cref{exp4}, our method covers a larger active area and is more expressive.

\section{Elaborate Version of Related works.}

\textbf{Human Motion Generation.}
Human motion generation can be broadly categorized into two primary approaches based on the type of input: 1) motion synthesis without conditions~\cite{raab2023modi, tevet2022human, Zhang_Black_Tang_2020, Zhao_Su_Ji_2020, Yan_Li_Xiong_Yan_Lin_2019,qin2024towards} and 2) motion synthesis with specified multimodal conditions, such as action labels~\cite{xu2023actformer, guo2020action2motion, petrovich2021action, lee2023multiact, dou2023c}, textual descriptions~\cite{wang2022humanise, chen2023executing,lin2018human,ahn2018text2action, tevet2022motionclip,  zhang2024motiondiffuse, tevet2022human,  petrovich2022temos, athanasiou2022teach,  lu2023humantomato,  guo2022generating,ahuja2019language2pose,qin2023motiontrack}, or audio and music~\cite{li2024lodge,tseng2023edge,Li_Zhao_Shi_Sheng,  Dabral_Mughal_Golyanik_Theobalt_2022,  li2021ai,  siyao2022bailando}.
With the widespread application of multimodal information~\cite{wang2025refdetector, wang2024referencing, wang2025mapping,wang2022cross, wang2024dual}, text-to-action conversion is currently one of the most important action generation tasks due to its user-friendly characteristics and the convenience of language input. Given the remarkable success of diffusion-based generative models in other domains~\cite{rombach2022high}, some approaches have employed conditional diffusion models for human motion generation~\cite{zhang2022motiondiffuse,zhang2023remodiffuse,zhang2024finemogen,chen2023executing}. Other works~\cite{zhang2023generating,guo2024momask,wang2023t2m} first discretize motions into tokens using vector quantization~\cite{van2017neural} and then predict the code sequence of motion.

\section{Metric Details and Further Results on Text-to-motion Generation.}

\begin{table}[ht]
\centering
\caption{Evaluation of text-to-motion capability on the HumanML3D test set, with comparison to state-of-the-art methods such as TM2T~\cite{guo2022tm2t}, T2M~\cite{guo2022generating}, MDM~\cite{shafir2023human}, MLD~\cite{chen2023executing}, MotionDiffuse~\cite{zhang2022motiondiffuse}, T2M-GPT~\cite{zhang2023generating}, and ReMoDiffuse~\cite{zhang2023remodiffuse}.}
\label{table3}
\vskip 0.15in
\begin{sc}
\resizebox{1\linewidth}{!}{
\begin{tabular}{lcccccc}
\hline\hline
\multirow{2}{*}{{Methods}} & \multicolumn{3}{c}{{R Precision$\uparrow$}}        & \multirow{2}{*}{{FID$\downarrow$}} & \multirow{2}{*}{{MultiModalit$\uparrow$}} & \multicolumn{1}{l}{\multirow{2}{*}{MultiModal Dist$\downarrow$}} \\ \cline{2-4}
                                  & {Top 1}    & {Top 2}    & {Top 3}    &                                           &                                                  & \multicolumn{1}{l}{}                                             \\ \hline
TM2T                              & \et{0.424}{.003}  & \et{0.618}{.003}  & \et{0.729}{.002}  & \et{1.501}{.017}                          & \et{2.424}{.093}                                 & \et{3.467}{.011}                                                 \\
T2M                               & \et{0.455}{.003}  & \et{0.636}{.003}  & \et{0.736}{.002}  & \et{1.087}{.021}                          & \et{2.219}{.074}                                 & \et{3.347}{.008}                                                 \\
MDM                               & -                 & -                 & \et{0.611}{.007}  & \et{0.544}{.044}                          & \etb{2.799}{.072}                                & \et{5.566}{.027}                                                 \\
MLD                               & \et{0.481}{.003}  & \et{0.673}{.003}  & \et{0.772}{.002}  & \et{0.473}{.013}                          & \et{2.413}{.079}                                 & \et{3.196}{.010}                                                 \\
MotionDiffuse                     & \et{0.491}{.001}  & \et{0.681}{.001}  & \et{0.782}{.001}  & \et{0.630}{.001}                          & \et{1.553}{.042}                                 & \et{3.113}{.001}                                                 \\
T2M-GPT                           & \et{0.492}{.003}  & \et{0.679}{.002}  & \et{0.775}{.002}  & \et{0.141}{.005}                          & \et{1.831}{.048}                                 & \et{3.121}{.009}                                                 \\
ReMoDiffuse                       & \etb{0.510}{.005} & \ets{0.698}{.006} & \etb{0.795}{.004} & \ets{0.103}{.004}                         & \et{1.795}{.043}                                 & \etb{2.974}{.006}                                                \\ \hline
Ours                              & \ets{0.505}{.002} & \etb{0.698}{.003} & \ets{0.794}{.002} & \etb{0.090}{.003}                         & \ets{2.563}{.071}                                & \ets{2.981}{.007}                                                \\ \hline\hline
\end{tabular}
}
\end{sc}
\vskip -0.1in
\end{table}
\textbf{Metric details}
For motion generation, we follow the common metrics of prior works~\cite{guo2022generating} to evaluate the text-to-motion generation performance. Global representations of motion and text descriptions are first extracted with the pre-trained network in~\cite{guo2022generating}, and then measured by the following five metrics:

\begin{itemize}
	\item R-Precision. Given one motion sequence and 32 text descriptions (1 ground-truth and 31 randomly selected mismatched descriptions), we rank the Euclidean distances between the motion and text embeddings. Top-1, Top-2, and Top-3 accuracy of motion-to-text retrieval are reported.
	\item Frechet Inception Distance (FID). We calculate the distribution distance between the generated and real motion using FID~\cite{heusel2017gans} on the extracted motion features.
        \item Multimodal Distance (MM-Dist). The average Euclidean distances between each text feature and the generated motion feature from this text.
        \item Multimodality (MModality). For one text description, we generate 30 motion sequences forming 10 pairs of motion. We extract motion features and compute the average Euclidean distances of the pairs. We finally report the average over all the text descriptions.
\end{itemize}

We present more metrics for comparison with other methods in~\Cref{table3}. In addition to the metrics discussed in the main text, we rank second in MM-Dist and MModality, respectively. This demonstrates great generative diversity and multimodal alignment capabilities.

\section{Introduction of HumanML3D.}

HumanML3D~\cite{guo2022generating} is currently the largest 3D human motion dataset with textual descriptions. The dataset contains 14,616 human motions and 44,970 text descriptions. The entire textual descriptions are composed of 5,371 distinct words. The motion sequences are originally from AMASS~\cite{mahmood2019amass} and HumanAct12~\cite{guo2020action2motion} but with specific pre-processing: motion is scaled to 20 FPS; those that are longer than 10 seconds are randomly cropped to 10-second ones; they are then re-targeted to a default human skeletal template and properly rotated to face Z+ direction initially. Each motion is paired with at least 3 precise textual descriptions. The average length of descriptions is approximately 12. According to~\cite{guo2022generating}, the dataset is split into training, validation, and test sets with proportions of 80\%, 5\%, and 15\%, respectively. We select the best FID model on the validation set and report its performance on the test set.

\section{User study on the token-to-pose translator}
To evaluate the effectiveness of token-to-pose translator, we conducted a blind user study with 10 participants. Using 10 reference images and 10 driving videos, we generated 20 clips. Participants compared video pairs for each input based on Motion Naturalness.
The use of the token-to-pose translator resulted in a 98\% win rate, demonstrating the effectiveness of this module in significantly reducing motion stiffness and enhancing the naturalness of body movements.
\begin{itemize}
\item Motion Naturalness: Participants compare the two generated videos and select the one in which the character's body movements appear less stiff, more natural, and exhibit greater detail and refinement in motion execution.
\end{itemize}

\begin{figure}[t]
\begin{center}
\centerline{\includegraphics[width=\columnwidth]{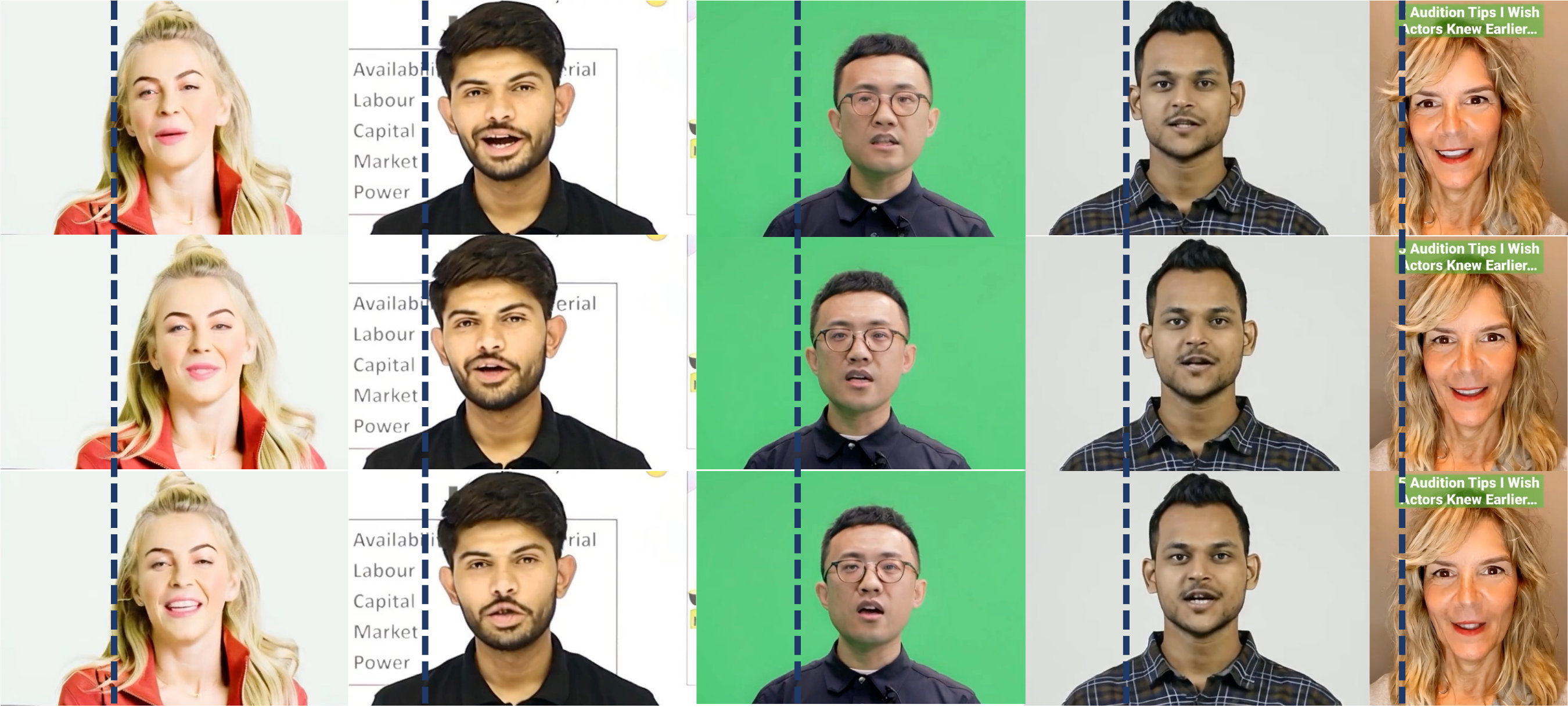}}
\caption{Visual illustration of natural head movement. }
\Description{}
\label{exp5}
\end{center}
\end{figure}

\begin{figure*}[ht]
\vskip 0.2in
\begin{center}
\centerline{\includegraphics[width=\textwidth]{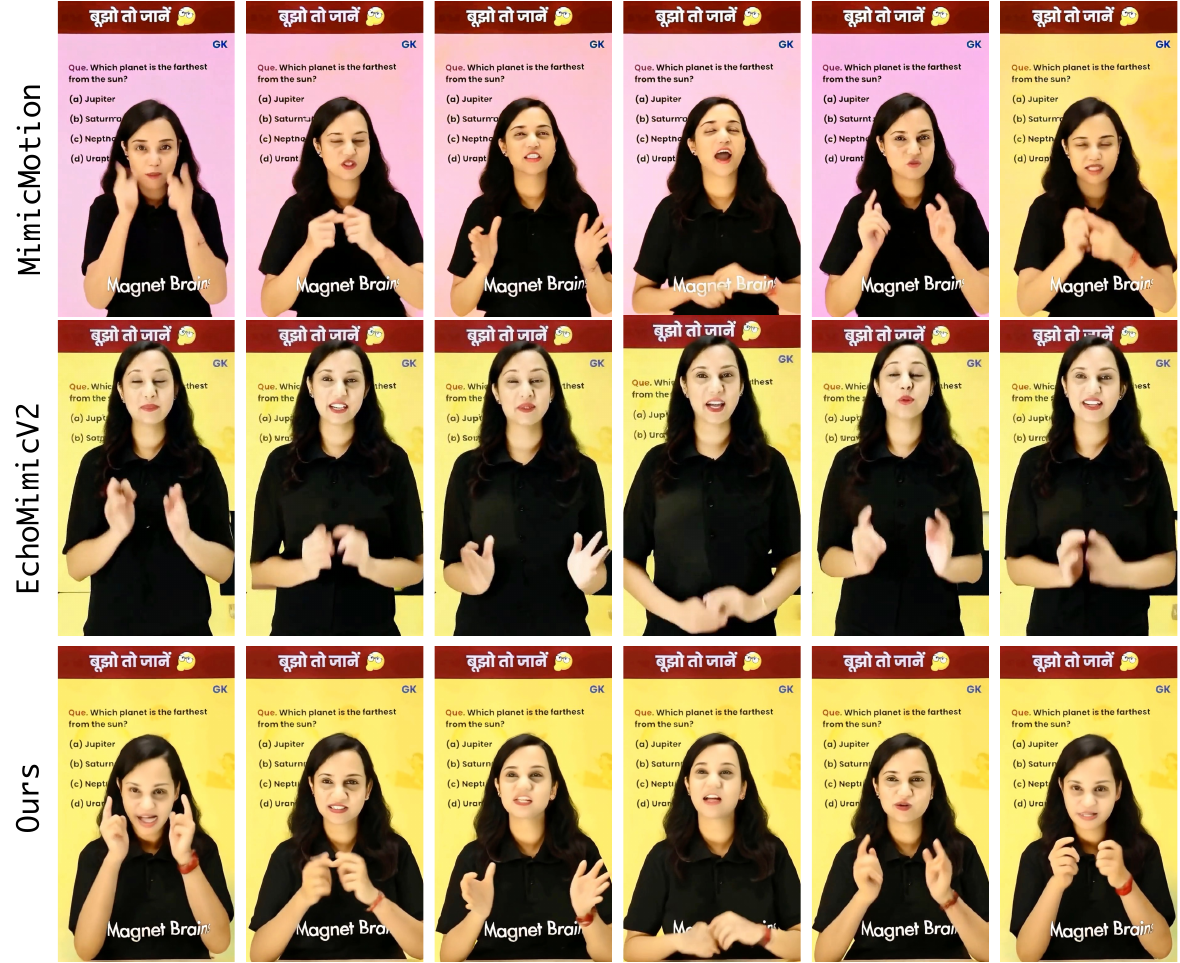}}
\caption{Comparisons with pose-driven body animation methods.}
\Description{}
\label{exp8}
\end{center}
\vskip -0.2in
\end{figure*}

\section{Visual illustration of natural head movement.}
The integration of 3D motion allows whole-body movement, with the head naturally following, aligning with real human speech. As shown in~\Cref{exp5}, our method produces natural head motion, enhancing realism and fluidity.

\begin{table}[t]
\centering
\caption{User study results on identity preservation~(IP), background preservation~(BP), temporal consistency~(TC) and visual quality~(VQ).}
\label{table3}
\resizebox{0.8\linewidth}{!}{
\begin{tabular}{lcccc}
\hline\hline
Method        & IP$\uparrow$ & BP$\uparrow$ & TC$\uparrow$ & VQ$\uparrow$ \\\hline
MimicMotion   & 30.5\%       & 16.5\%       & 13.5\%          & 33.0\%          \\
EchoMimicV2   & 23.0\%       & 38.0\%       & 47.0\%          & 23.5\%          \\
VersaAnimator & 96.5\%       & 95.5\%       & 89.5\%          & 93.5\%          \\ \hline\hline
\end{tabular}
}
\end{table}

\section{User study in comparison with SOTA methods}
To evaluate our method against state-of-the-art approaches, we conducted a blind user study with 10 participants. Using 10 reference images and 10 driving videos, we generated 30 clips across three methods. Participants compared video pairs for each input based on visual quality, identity preservation, background preservation, and temporal consistency. Each comparison is repeated $C_{2}^{3}$ times. As shown in~\Cref{table3}, our method consistently outperformed others across all criteria. 

To ensure the feedback reflects practical applicability, the ten participants in our user study come from diverse academic backgrounds. Since many of them do not specialize in computer vision, we provided detailed explanations for each evaluation criterion to assist their judgments:
\begin{itemize}
\item Identity Preservation: Participants compare the reference image with the two generated videos and determine which video’s character more closely resembles the person in the reference image.
\item Temporal Consistency: Participants observe the motion dynamics of the character within each video and assess which one displays smoother and more coherent movement over time.
\item Visual Quality: This criterion involves a more subjective assessment. Participants are asked to evaluate the overall visual fidelity, taking into account factors such as artifacts (e.g., flickering, distortions, afterimages), motion realism (e.g., smoothness, physical plausibility), and the general believability of the animation.
\item Background Preservation: Participants compare the reference image with the two generated videos and evaluate which video maintains greater consistency in the background environment, including aspects such as spatial layout, lighting conditions, and object appearances.
\end{itemize}

\section{Comparisons with Pose-driven Body Animation Methods.}

We present more comparisons results with pose-driven body animation methods.
The visual results in \Cref{exp8} demonstrate that VersaAnimator maintains superior structural integrity and identity consistency in local regions, such as the hands and face, when compared to the current state-of-the-art methods. 

\section{Video demos.}
The folder contains demo videos showcasing speakers of various nationalities and languages, along with cross-nationality generalization results, where a single reference image is driven by audio from speakers of different linguistic backgrounds.

\balance

\end{document}